\documentclass{article}

\usepackage{graphicx}
\usepackage{color}

\usepackage[nonatbib, final]{nips_2018}

\usepackage[utf8]{inputenc} 
\usepackage[T1]{fontenc}    
\usepackage{hyperref}       
\usepackage{url}            
\usepackage{booktabs}       
\usepackage{amsfonts}       
\usepackage{nicefrac}       
\usepackage{microtype}      
\usepackage{authblk}
\makeatletter
\renewcommand\AB@affilsepx{, \protect\Affilfont}
\makeatother

\title{Policy Gradient Search:  Online Planning \\and Expert Iteration without Search Trees}

\author[1]{\textbf{Thomas Anthony}}
\author[2,3]{\textbf{Robert Nishihara}}
\author[2,3]{\textbf{Philipp Moritz}}
\author[4]{\\\textbf{Tim Salimans}}
\author[2]{\textbf{John Schulman}\hspace{9pt}\thanks{Corresponding author: Thomas Anthony <twa@google.com>. Work conducted at OpenAI}\hspace{-9pt}}
\affil[1]{DeepMind}
\affil[2]{OpenAI}
\affil[3]{University of California, Berkeley}
\affil[4]{Google Brain, Amsterdam}

\begin{document}

\maketitle

\begin{abstract}
Monte Carlo Tree Search (MCTS) algorithms perform simulation-based search to improve policies online. During search, the simulation policy is adapted to explore the most promising lines of play. MCTS has been used by state-of-the-art programs for many problems, however a disadvantage to MCTS is that it estimates the values of states with Monte Carlo averages, stored in a search tree; this does not scale to games with very high branching factors. We propose an alternative simulation-based search method, Policy Gradient Search (PGS), which adapts a neural network simulation policy online via policy gradient updates, avoiding the need for a search tree. In Hex, PGS achieves comparable performance to MCTS, and an agent trained using Expert Iteration with PGS was able defeat MoHex 2.0, the strongest open-source Hex agent, in 9x9 Hex. 
\end{abstract}

\section{Introduction}

The value of Monte Carlo Tree Search (MCTS) \cite{kocsis2006bandit, browne2012survey, coulom2006efficient} for achieving maximal test-time performance in games such as Go and Hex has long been known. More recent works have also shown that incorporating planning into the training of reinforcement learning (RL) agents with Expert Iteration (\textsc{ExIt}) \cite{silver2017mastering, anthony2017thinking, Silver18AlphaZero} allows a pure RL approach to achieve state-of-the-art performance \textit{tabula rasa} in many classical board games.

However, MCTS builds an explicit search tree, storing visit counts and value estimates at each node - in other words, creating a tabular value function. To be effective, this requires that nodes in the search tree are visited multiple times. This is true in many classical board games, but many real world problems have large branching factors that make MCTS hard to use. Large branching factors can be caused by very large action spaces, or chance nodes. In the case of large action spaces, a prior policy can be used to discount weak actions, reducing the effective branching factor. Stochastic transitions are harder to deal with, as prior policies cannot be used to reduce the branching factor at chance nodes.

In contrast, Monte Carlo Search (MCS) \cite{tesauro1997line} algorithms have no such requirement. Whereas MCTS uses value estimates in each node to adapt the simulation policy, MCS algorithms have a fixed simulation policy throughout the search. However, because MCS does not improve the quality of simulations during search, it produces significantly weaker play than MCTS.

To adapt simulation policies in problems where we don't visit states multiple times, we can search over a restricted version of the problem, where multiple visits to states do occur \cite{couetoux2011continuous}, or we can generalise knowledge between different states during search. We take the latter approach. To this end, we propose Policy Gradient Search (PGS) a search algorithm which trains its simulation policy during search using policy gradient RL. This gives the advantages of an adaptive simulation policy, without requiring an explicit search tree to be built. 

We test PGS on 9x9 and 13x13 Hex, a domain where MCTS has been used in all state-of-the-art players since 2009. We find that PGS is significantly stronger than MCS, and competitive with MCTS. Additionally, we show that Policy Gradient Search Expert Iteration is able to defeat \textsc{MoHex 2.0} in 9x9 Hex \textit{tabula rasa}, the first agent to do so without using explicit search trees. 

Section \ref{background} covers background material, and section \ref{pgs} describes the PGS algorithm. In section \ref{exp} we assess the strength of PGS as a test-time decision maker, while in section \ref{pgsexit} we show results from using PGS within \textsc{ExIt}. Sections \ref{related} and \ref{discuss} discuss connections to previous works and future directions.

\section{Background}
\label{background}

\subsection{Markov Decision Processes}

We consider sequential decision making in a Markov Decision Process (MDP)\cite{sutton2018reinforcement}. At each timestep $t$, an agent observes a state $s_t$ and chooses an action $a_t$ to take. In a terminal state $s_T$, an episodic reward $R$ is observed, which we intend to maximise. We can easily extend to two-player, perfect information, zero-sum games by learning policies for both players simultaneously, which aim to maximise the reward for the respective player, so we will refer to both MDPs and two-player, perfect information, zero-sum games as MDPs throughout this work.

We call a distribution over the actions $a$ available in state $s$ a \textit{policy}, and denote it $\pi(a|s)$. The value function $V^\pi(s)$ is the mean reward from following $\pi$ starting in state $s$. By $Q^\pi(s,a)$ we mean the expected reward from taking action $a$ in state $s$, and following policy $\pi$ thereafter.

\subsection{Hex}

\begin{figure}[h!]
	\begin{center}
		\includegraphics[scale=0.3]{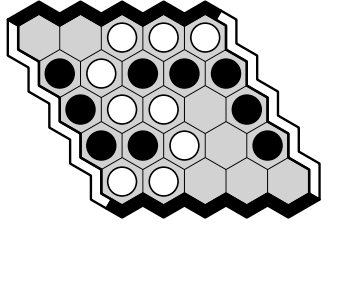}\\
		\vspace*{-8mm}
		\caption{A $5\times5$ Hex game, won by white. Figure from Huang et al. \cite{huang2013mohex}.}
		\label{hex}
	\end{center}
\end{figure}

Hex is a two-player connection-based game played on an $n \times n$ hexagonal grid. The players, denoted by colours black and white, alternate placing stones of their colour in empty cells. The black player wins if there is a sequence of adjacent black stones connecting the North edge of the board to the South edge. White wins if they achieve a sequence of adjacent white stones running from the West edge to the East edge. (See figure \ref{hex}).

Hex requires complex strategy, making it challenging for deep RL algorithms; its large action set and connection-based rules means it shares similar challenges for AI to Go. However, games can be simulated efficiently because the win condition is mutually exclusive (e.g. if black has a winning path, white cannot have one), its rules are simple, and permutations of move order are irrelevant to the outcome of a game. These properties make it an ideal test-bed for RL.

\subsection{Monte Carlo Tree Search}

Monte Carlo Tree Search (MCTS) is an any-time best-first tree-search algorithm. It uses repeated game simulations to estimate the value of states, and expands the tree further in more promising lines. When all simulations are complete, the most explored move is taken.

Each node of the search tree corresponds to a possible state $s$ in the game. The root node corresponds to the current state, its children correspond to the states resulting from a single move from the current state, etc. The edge from state $s_1$ to $s_2$ represents the action $a$ taken in $s_1$ to reach $s_2$, and is identified by the pair $(s_1,a)$.

At each node we store $n(s)$, the number of iterations in which the node has been visited so far. Each edge stores both $n(s,a)$, the number of times it has been traversed, and $r(s,a)$ the sum of all rewards obtained in simulations that passed through the edge, so $Q(s,a) = r(s,a)/n(s,a)$ is the Monte Carlo estimate of the action-value.

The simulation policy depends on these statistics, as well as prior information in the form of a policy learnt offline. The most commonly used simulation policies are variants of the UCT formula \cite{kocsis2006bandit}, which trades exploration and exploitation within the tree search.

When an action $a$ in a state $s_L$ is chosen that takes us to a position $s'$ not yet in the search tree, we perform an expansion, adding $s'$ to the tree as a child of $s_L$. We estimate the value of the state $s'$, either with a Monte Carlo rollout following some \textit{default policy}, or, in more recent works \cite{silver2016mastering, silver2017mastering, anthony2017thinking}, with a neural-network value estimate.
This reward signal is propagated through the tree (a \textit{backup}), with each node and edge updating statistics for visit counts $n(s)$, $n(s,a)$ and total returns $r(s,a)$.

\subsection{Monte Carlo Search}
\label{mcs}

Monte Carlo Search (MCS) is a simpler search algorithm than MCTS. The goal is, given a state $s_0$ and policy $\pi$, to evaluate $Q^\pi(s_0,a)$. This is done by repeatedly simulating episodes starting from $s_0$ according to $\pi$, the mean outcome is then used to estimate the $Q$-values.

When used to select an action at $s_0$, the first action from $s_0$ can be chosen from a policy other than $\pi$. By following a bandit algorithm such as UCB \cite{auer2002using} we can direct more resources to simulating actions that are likely to have the highest $Q$-value, resulting in a more efficient search. This is similar to the use of UCT in MCTS, but only applied to the first action in each simulation.

When an accurate value function is available, we can stop simulations before the end of the episode, and bootstrap from the estimated value of the state at which we stopped. This is known as \textit{truncated} Monte Carlo simulation.

Compared to MCTS, MCS does not adapt its simulation policy. We can convert an MCTS algorthm to a very similar MCS algorithm by, at every node except the root node $s_0$, replacing the adaptive UCT formula with a policy that samples from the original policy $\pi$.

\subsection{Expert Iteration}

Search algorithms plan strong actions from a single state $s_0$, but do not learn information that generalises to different positions. In contrast, deep neural networks are able to generalise knowledge across a state space.

Expert Iteration (\textsc{ExIt}) \cite{anthony2017thinking} algorithms combine search-based planning with deep learning. A planning algorithm, referred to as the expert, is used to discover improvements to the current policy. A neural network acts as an apprentice, imitating the expert policy and estimating the value function. The planning algorithm can use the neural network policy and value estimates to improve the quality of its plans, resulting in a cycle of mutual improvement. This is a version of \textit{Approximate Policy Iteration} (API)\cite{scherrer2015approximate, lagoudakis2003reinforcement}, where the policy improvement operator performs a multiple-step policy improvement, rather than a 1-step greedy improvement.

AlphaZero \cite{silver2017mastering,Silver18AlphaZero} also uses MCTS as a multi-step policy improvement operator, training residual neural networks to predict the policy and value of self-play games of the MCTS. It achieved state of the art, superhuman play in Chess, Go and Shogi.

\section{Policy Gradient Search}
\label{pgs}

Policy Gradient Search works by applying a model-free RL algorithm to adapt the simulations in Monte Carlo Search. We will assume that a prior policy $\pi$ and a prior value function $V$ are provided, which have been trained on the full MDP.

The algorithm must represent everything it learns through non-tabular function approximators, otherwise it will suffer the same drawbacks as MCTS. MCTS is already a form of self-play RL, however we cannot directly adapt it to use function approximation, because UCT formulae rely on count-based exploration rules.

Instead, we use policy gradient RL to train the simulation policy. Our simulation policy $\pi_{sim}$ is represented by a neural network with identical architecture to the global policy network. At the start of each game, the parameters of the policy network are set to those of the global policy network.

Because evaluating our simulation policy is expensive, we do not simulate to a terminal state, but instead use truncated Monte Carlo simulation. Choosing when to truncate a simulation is not necessarily simple, the best choice may depend on the MDP itself. If simulations are too short, they may fail to contain new information, or not give a long enough horizon search. Too long simulations will be wasteful.

Reasonable strategies could include simulating for a fixed horizon $H$; or for a gradually increasing horizon $H(n)$ on the $n$th simulation; or simulating until a threshold on the probability density of the action sequence is reached, to induce deeper search down the principle variation. For Hex, we use the same strategy as employed by MCTS algorithms: running each simulation until the action sequence of the simulation is unique.\footnote{Noting that an alternative would be needed in domains with very large action spaces}

Once we reach a final state for the simulation $s_L$ after $t$ steps, we estimate the value of this state using the global value network $V$, and use this estimate to update the simulation policy parameters $\theta$ using \textsc{Reinforce} \cite{williams1992simple}:
$$\theta = \theta + \alpha V(s_L)\sum_{i=1}^t\nabla_\theta\log{\pi(a_i|s_i)}$$
Where $\alpha$ is a learning rate. In Hex, values are scaled between -1 and 1, for other problems, a non-zero baseline may be necessary. These updates can be seen as fine-tuning the global policy to the current sub-game.

Because the root node is visited in every simulation, as with MCS, we can use a bandit-based approach to select the first action $a_0$ of each simulation. We adopt the PUCT formula \cite{silver2017mastering, rosin2011multi} for this, greedily choosing the action that maximises:
$$PUCT(s_0,a) = Q(s_0,a) + c_{puct}\pi(s_0,a)\frac{\sqrt{n(s_0)}}{1+n(s_0,a)}$$
Where $c_{puct}$ is a hyperparameter. $Q(s_0,a)$ is the average return from all simulations so far that started with the action $a$. $\pi(s_0,a)$ is the original global policy at $s_0$. Every subsequent action of the simulation $a_{1:t}$ is sampled from $\pi_{sim}$.

\subsection{Parameter Freezing during Online Adaptation}
\label{eff}

During testing, online search algorithms are usually used under a time constraint, so, compared to standard RL problems, orders of magnitude fewer simulations will be used. It is also important to ensure that our algorithm does not require too much computation per simulation step. When used for offline training in Expert Iteration, the efficiency of the search method is still crucial: too slow, and it would be more efficient to use a worse but faster planner, and run for a greater number of iterations.

We use a residual neural network with the architecture introduced by Silver et al.\cite{silver2017mastering}, with 19 residual blocks and separate policy and value heads. In order to learn a policy effective across the entire state space from a dataset of millions of positions, the global neural network is very large and expensive to evaluate. Far fewer parameters are sufficient for the online adaptation, or even preferable if it regularises the fine-tuning.

PGS is more expensive than MCS because it must perform the policy gradient update to the neural network parameters. The backward pass through our network takes approximately twice as long as the forward pass, making PGS 3-4 times more expensive than MCS. In order to reduce the computational cost of the algorithm, during policy gradient search, we adapt the parameters of the policy head only. This reduces the flops used by the backward pass by a factor of over 100, making the difference in computational cost between MCS and PGS negligible.

(In games such as Hex where states are visited multiple times, an additional optimisation can then be made: the forward pass through the fixed part of the network can be cached, rather than being recalculated for every visit of each simulation. This is similar to storing the prior policy at each node of MCTS, and substantially reduced the runtime of our experiments.)

\subsection{Note on Batch Normalisation}

Our neural network uses batch normalisation \cite{ioffe2015batch}. In all instances, the global neural networks have been trained on datasets of states from many independently sampled games of Hex.

During search, the input distribution is substantially changed to consist of many highly correlated states. Using mini-batch statistics for batch normalisation therefore results in a large shift in the policy. So during PGS we freeze the parameters for batch normalisation, and calculate the normalisation using population rather than mini-batch statistics, as is usual for inference.

\section{Policy Gradient Search as an Online Planner}
\label{exp}

We evaluate Policy Gradient Search on the game of Hex. Hex has a moderate branching factor and deterministic transitions, meaning MCTS is very effective in this domain, this allows us to directly compare the strength of PGS to MCTS. As noted in section \ref{eff}, faster experimentation is possible in such domains, too. In this section we measure the performance of PGS for maximising agent performance at test time. 

\subsection{Baselines}

In our experiments we use two baseline search algorithms: MCTS and MCS. MCTS provides a strong baseline, but becomes harder to use in some MDPs with very large branching factors. In contrast, MCS is easier to apply to general MDPs, but is weaker than MCTS in Hex.

Our MCTS algorithm is the same as used by AlphaZero\cite{Silver18AlphaZero}. That is, we use the PUCT formula for our tree policy, $c_{puct} = 5$, and use a value network for leaf evaluations. Our MCS is as described in section \ref{mcs}, the same as the MCTS but sampling from the prior policy instead of using PUCT at every node except the root. It is therefore also the same as our PGS algorithm with a learning rate of 0.

For all search algorithms, simulations are completed in batches of 32, with virtual losses added wherever the PUCT formula was used to encourage diversity in the simulations \cite{segal2010scalability}.

\subsection{Description of the Neural Networks}
\label{nets}

To show general applicability, we test using multiple different global neural networks, trained with different variants of Expert Iteration. In each case, we tune the PGS learning rate before testing, and compare PGS to MCS and MCTS using the same neural network. The networks used are:

\begin{enumerate}
    \item A residual neural network for 9x9 Hex trained on the dataset generated in the distributed training run from Anthony et al. \cite{anthony2017thinking}
    \item The network at the end of training for our version of AlphaZero (see section \ref{pgsexit}) applied to 9x9 Hex
    \item A network from early in training with Policy Gradient Search Expert Iteration (\textsc{PGS-ExIt}, section \ref{pgsexit}), applied to 9x9 Hex i.e. at epoch 10 of 450
    \item A network from approximately half way through training with \textsc{PGS-ExIt} on 9x9 Hex, i.e. at epoch 230 of 450
    \item The network at the end of training with \textsc{PGS-ExIt} applied to 9x9 Hex, i.e. at epoch 450 of 450
    \item The network at the end of training with our version of AlphaZero, applied to 13x13 Hex
\end{enumerate}

\subsection{Results}

For each neural network, we ran a round-robin tournament between the raw neural network and four search algorithms MCS, MCTS, PGS, and PGS without parameter freezing (PGS-UF). To overcome Hex's first player advantage, each pair of agents played $2n^2$ games against each other, one per colour per legal first more. Note that this compresses the Elo scale compared to tournament play, as many legal opening moves give one or other player a significant advantage.

Each searcher used 800 search iterations per move, with no pondering between moves. Elo ratings were calculated using BayesElo \cite{bayeselo}, with the scale shifted so the mean estimate for the raw neural network's Elo was 0 in each case, giving a different Elo scale for each tournament. Results are presented in the table below, along with the values of $\alpha$ used in PGS.

\begin{center}
	\begin{tabular}{ c|c|c|c|c c|c c}
		\toprule
		NN & NN Elo & MCS Elo & MCTS Elo & PGS Elo & PGS $\alpha$ & PGS-UF Elo & PGS-UF $\alpha$\\
		\midrule
		1 & $0 \pm 46$ & $354 \pm 30$ & $\mathbf{462 \pm 31}$ & $\mathbf{430 \pm 30}$ & 5e-4 & $349 \pm 30$ & 1e-6\\ 
		2 & $0 \pm 28$ & $\mathbf{96 \pm 27}$ & $\mathbf{123 \pm 27}$ & $\mathbf{117 \pm 27}$ & 1e-4 & $\mathbf{94 \pm 27}$ & 5e-6\\ 
		3 & $0 \pm 46$ & $335 \pm 30$ & $\mathbf{481 \pm 31}$ & $\mathbf{466 \pm 30}$ & 1e-4 & $380 \pm 30$ & 5e-5\\ 
		4 & $0 \pm 30$ & $153 \pm 27$ & $\mathbf{244 \pm 28}$ & $\mathbf{198 \pm 28}$ & 1e-4 & $\mathbf{179 \pm 27}$ & 5e-6\\ 
		5 & $0 \pm 29$ & $125 \pm 27$ & $\mathbf{171 \pm 27}$ & $\mathbf{166 \pm 27}$ & 1e-4 & $\mathbf{138 \pm 27}$ & 5e-6\\ 
		6 & $0 \pm 25$ & $161 \pm 25$ & $\mathbf{269 \pm 23}$ & $\mathbf{239 \pm 23}$ & 2e-5 & - & - \\ 
		\bottomrule
	\end{tabular}
\end{center}

In all cases, all search algorithms significantly outperformed the raw neural network. Most differences between the search algorithms are smaller, but overall trends are clear: on average PGS is $\sim65$ Elo stronger than MCS, and MCTS is $\sim20$ Elo stronger than PGS. PGS without freezing parameters (PGS-UF) was found to be weaker than PGS, even disregarding the additional computational cost.

\begin{figure}[h!]
	\begin{center}
		\includegraphics[width=\textwidth]{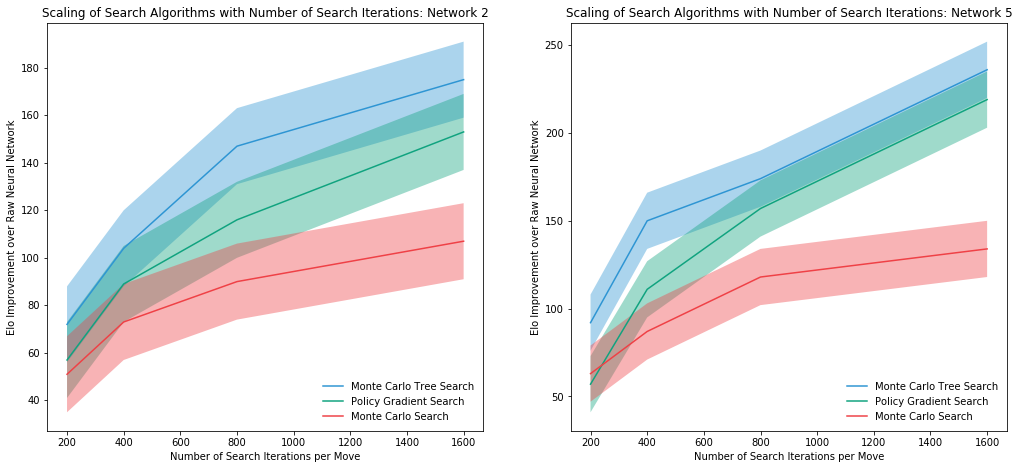}\\
		\caption{Graphs showing how the strength of search algorithms scales with the number of search iterations. Shaded regions represent $95\%$ confidence intervals}
		\label{scaling_fig}
	\end{center}
\end{figure}

We also tested how the performance of the different search algorithms scales with different numbers of search iterations, in a range from 200 to 1600 search iterations per move, our results are plotted in figure \ref{scaling_fig}. In both cases we see that the adaptive search algorithms, PGS and MCTS, scale much more effectively with number of search iterations than does MCS. 

PGS might scale less well than MCTS if the capacity for the policy head to represent adaptations were saturated. We find no evidence that this occurs, but note that 1600 iterations per move is still a fairly short search, such an effect may still take place in longer searches.

\section{Policy Gradient Search Expert Iteration}
\label{pgsexit}
One motivation of this work is the value of online planning algorithms during training of RL agents. To this end, we used PGS as an expert in \textsc{ExIt}, comparing again to the baseline MCS and MCTS agents.

\subsection{Expert Iteration Setup}

We closely follow the self-play and distillation schemes of AlphaZero \cite{silver2017chess}, which we summarise here. Data is generated via self-play of the expert search algorithm on multiple workers. Whenever a game is finished, all states $s_i$ from the game are added to a replay buffer, with the game result $z$ and expert search policy $p(s_i, a) = n(s_i,a)/n(s_i)$ for each state. Asynchronously, the neural network is trained on the data in the replay buffer. \footnote{Asynchronous training and data generation was implemented with Ray \cite{moritz2017ray}}

For the first 30 moves of each self-play game, the expert takes actions according to the search distribution, i.e. $\pi_{expert}(s, a) = n(s,a)/n(s)$. Thereafter actions are chosen greedily. At the root node, Dirichlet noise is added to the prior policy in the PUCT formula: $\pi'(s_0, a) = 0.75\pi(s_0, a) + 0.25\eta$, $\eta \sim Dir(0.12)$. In $90\%$ of games, resignation is used: if the average return of search simulations from the current state is below a resignation threshold, the game is resigned. In the other $10\%$ of games no resignation is used, these games are used to calculate a threshold with a false positive rate below $5\%$.

The replay buffer stored the 10,000,000 most recent states. The neural network was trained with a batch size of 1024, optimised with fixed learning rate of 0.01. We used momentum with a momentum parameter of 0.9. A cross-entropy loss was used for optimising the policy, mean square error for optimising the value function, and an L2 weight regulariser with weight $10^{-4}$ was used.

For all search algorithms, $c_{puct} = 5$ wherever the PUCT formula is used. For \textsc{PGS-ExIt}, we used an `inner' learning rate during PGS of $\alpha = $ 5e-4. 

\subsection{Results}

\begin{figure}[h!]
	\begin{center}
		\includegraphics[width=\textwidth]{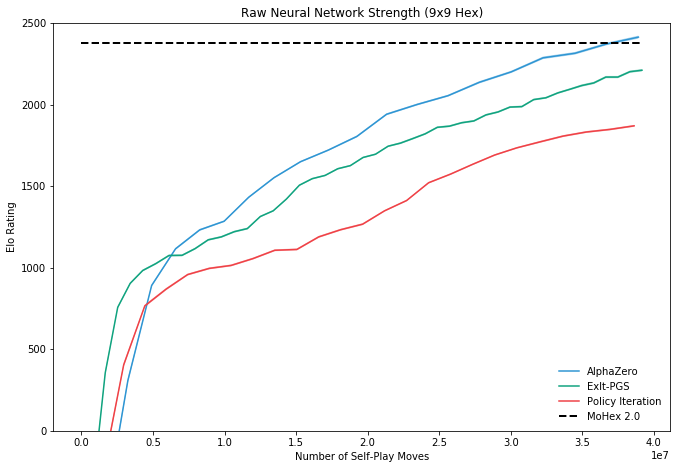}\\
		\caption{Strength of the raw neural networks throughout training, calculated by a round-robin tournament between the networks.}
		\label{graph}
	\end{center}
\end{figure}

Our results are in line with those from section \ref{exp}, with PGS performing better than MCS, but not as well as MCTS. Over the course of training, the differences in strength of the agents has compounded through repeated application of better or worse experts. AlphaZero (i.e. \textsc{MCTS-ExIt}) significantly outperforms \textsc{PGS-ExIt}, which in turn significantly outperforms Approximate Policy Iteration (i.e. \textsc{MCS-ExIt}). We show the strength of the raw policy networks (with no search) throughout training in figure \ref{graph}.

The `inner' learning rate $\alpha$ for \textsc{PGS-ExIt} was chosen based on the optimum value for Network 1 from section \ref{nets}. Subsequent tests showed this not to be optimal for networks trained by \textsc{PGS-ExIt}. Indeed, the benefit of using PGS over MCS is much reduced by this sub-optimal $\alpha$. Presumably, with a better setting for this parameter, the performance of \textsc{PGS-ExIt} could be improved. Over the course of training, approximately 2 million games were played. In contrast, tuning this hyperparameter requires 2000 games; automatic tuning would not significantly increase the cost of the algorithm.

\subsection{Comparison to \textsc{MoHex}}
\label{mohex}

\textsc{MoHex 2.0} is the strongest open-source Hex agent.\footnote{More recently, stronger agents have been published, but are not available for benchmarking. 9x9 is a smaller boardsize than is used for tournament play, and has been weakly solved \cite{pawlewicz2013scalable}.} It is a classical MCTS program with many Hex specific improvements, including an end-game solver, virtual connection calculator, and pattern based rollouts. In contrast, all agents in this paper were trained \textit{tabula rasa}. We played a head-to-head match between \textsc{MoHex 2.0}, with 10,000 iterations, and \textsc{PGS-ExIt} with 800 iterations. Playing 4 games from each first move with each colour, \textsc{PGS-ExIt} won by $375$ games to $273$, 55 Elo stronger. The final agent trained with Policy Iteration lost by $540$ games to $108$, a gap of 280 Elo, though both API and \textsc{PGS-ExIt} were still improving at the end of training.

All previous competitive Hex agents have used explicit tree search algorithms at test time, and many also use them during training \cite{anthony2017thinking}, did not learn to play tabula rasa \cite{arneson2010hex, huang2013mohex, takada2017reinforcement}, or both \cite{gao2017move}; we present here the first competitive agents that entirely forgo both tree search and prior Hex knowledge.

\section{Related Work}
\label{related}

Temporal Difference Search (TDS) \cite{silver2009reinforcement} is another search algorithm that employs a model-free RL algorithm for online search. Like PGS, it doesn't require an explicit search tree because it uses function approximators. Because TDS defines the simulation policy with value functions, it requires many more function evaluations per move than does PGS- this is okay with linear function approximators, as used by Silver \cite{silver2009reinforcement}, but would not be feasible with the large neural networks used in this work. Graf and Platzner \cite{graf2015adaptive} use policy gradient updates to improve a linear default policy of a classical MCTS Go program, this can be seen as a combination of PGS and MCTS. They show that this combination results in stronger play than applying MCTS with a fixed default policy.

\textsc{PGS-ExIt} decomposes the overall RL problem into many sub-problems, one per self-play game, and attempts to solve (or at least make progress on) each of the sub-problems with a model-free RL algorithm. The solutions to sub-problems are distilled back into a global network. Recent works on multi-task RL, such as Distral \cite{teh2017distral}, follow a similar pattern. Divide and Conquer RL \cite{ghosh2017divide} also attempts to solve a single MDP by considering it as multiple different sub-problems, the division in Divide and Conquer RL is based on auxiliary context information, which determines which of a small set of sub-problems any given episode belongs to. In \textsc{PGS-ExIt}, every episode effectively belongs to its own unique context, a more rudimentary technique which nonetheless has the advantage of requiring no additional information.

Previous works have applied model-free Reinforcement Learning algorithms to train networks to play Hex previously, albeit never trained tabular rasa. NeuroHex \cite{young2016neurohex} used deep Q-learning, while Gao et al.\cite{gao2018adversarial} proposed several variants of policy gradient algorithms for alternating move games. These networks are strong: they can win some games against MoHex 2.0 without test-time tree search, but they remain significantly weaker than MoHex. Gao et al.\cite{gao2018adversarial} also showed that combining their neural network with MoHex 2.0 resulted in an tree search algorithm stronger than MoHex 2.0.

\section{Discussion and Future Work}
\label{discuss}

In this work, we have presented Policy Gradient Search, a search algorithm for online planning that does not require an explicit search tree. We have shown that PGS is an effective planning algorithm. In our tests, it was slightly weaker than, but competitive with, MCTS, while significantly outperforming MCS for test-time decision making, in both 9x9 and 13x13 Hex. 

PGS is also effective during training when used within the Expert Iteration framework, resulting in the first competitive Hex agent trained tabula rasa without use of a search tree. In contrast, similar \textsc{Reinforce} algorithm alone was previously been found to not be competitve with an \textsc{ExIt} algorithm that used MCTS experts. \cite{anthony2017thinking}

Ablations show that \textsc{PGS-ExIt}, significantly outperforms MCS in the Expert Iteration framework, and also provide the first empirical data showing that \textsc{MCTS-ExIt} algorithms outperform traditional policy iteration approaches.

The results presented in this work are on the deterministic, discrete action space domain of Hex. This allowed for direct comparison to MCTS, but the most exciting potential applications of PGS are to problems where MCTS cannot be readily used, such as problems with stochastic state transitions or continuous action spaces. We leave extending PGS and \textsc{PGS-ExIt} to such domains to future work.

The implementation of PGS presented in this work is in some ways rudimentary, using vanilla REINFORCE with stochastic gradient descent. Policy gradient algorithms for model-free RL have benefited from the use of more advanced optimisation algorithms such as ADAM \cite{kingma2014adam}, and enhancements such as PPO \cite{schulman2017proximal}. Similar techniques might also improve the performance of PGS.

\bibliographystyle{unsrt}
\bibliography{bibliography}

\end{document}